\documentclass[11pt,a4paper]{article}

\usepackage[margin=1in]{geometry}
\usepackage{amsmath,amssymb}
\usepackage{graphicx}
\usepackage{booktabs}
\usepackage{array}
\usepackage{multirow}
\usepackage[table]{xcolor}
\usepackage{hyperref}
\usepackage{url}
\usepackage{natbib}
\usepackage{setspace}
\usepackage{caption}
\usepackage{subcaption}
\usepackage{enumitem}
\usepackage{mdframed}
\usepackage{titlesec}
\usepackage{fancyhdr}
\usepackage{abstract}
\usepackage[T1]{fontenc}

\definecolor{darkblue}{HTML}{1F3864}
\definecolor{midblue}{HTML}{2E4F8A}
\definecolor{rowgray}{HTML}{F4F6F9}
\definecolor{headgray}{HTML}{EEEEEE}

\hypersetup{
    colorlinks=true,
    linkcolor=midblue,
    citecolor=midblue,
    urlcolor=midblue,
    pdftitle={Self-Anchoring Calibration Drift in Large Language Models},
    pdfauthor={Harshavardhan},
}

\titleformat{\section}{\large\bfseries\color{darkblue}}{\thesection}{1em}{}
\titleformat{\subsection}{\normalsize\bfseries\color{midblue}}{\thesubsection}{1em}{}
\titleformat{\subsubsection}{\normalsize\bfseries\color{midblue}}{\thesubsubsection}{1em}{}
\titlespacing*{\section}{0pt}{14pt plus 2pt minus 2pt}{6pt plus 1pt}
\titlespacing*{\subsection}{0pt}{10pt plus 2pt minus 2pt}{4pt plus 1pt}
\titlespacing*{\subsubsection}{0pt}{8pt plus 2pt minus 2pt}{3pt plus 1pt}

\newcolumntype{L}[1]{>{\raggedright\arraybackslash}p{#1}}
\newcolumntype{C}[1]{>{\centering\arraybackslash}p{#1}}
\rowcolors{2}{rowgray}{white}

\newmdenv[
  leftline=true, rightline=false, topline=false, bottomline=false,
  linewidth=3pt, linecolor=midblue,
  innerleftmargin=10pt, innerrightmargin=10pt,
  innertopmargin=4pt, innerbottommargin=4pt,
  backgroundcolor=white
]{hypbox}

\begin{document}
\sloppy

% ── Title block ───────────────────────────────────────────────────────────────
\begin{center}
  {\LARGE\bfseries\color{darkblue}
    Self-Anchoring Calibration Drift in Large Language Models:\\[0.4em]
    How Multi-Turn Conversations Reshape Model Confidence
  \par}
  \vspace{0.8em}
  {\large Harshavardhan\par}
  \vspace{0.2em}
  {\normalsize Independent Researcher\quad
    \href{mailto:harsh@link.cuhk.edu.hk}{\texttt{harsh@link.cuhk.edu.hk}}\par}
  \vspace{0.2em}
  {\normalsize Code \& data: \href{https://github.com/hvardhan878/calibration-drift}{\texttt{github.com/hvardhan878/calibration-drift}}\par}
  \vspace{0.2em}
  {\normalsize February 2026\par}
\end{center}

\vspace{0.5em}

\begin{abstract}
  \noindent
  We introduce \textbf{Self-Anchoring Calibration Drift (SACD)}, a hypothesized tendency for large language models (LLMs) to show systematic changes in expressed confidence when building iteratively on their own prior outputs across multi-turn conversations. We report an empirical study comparing three frontier models — Claude Sonnet~4.6, Gemini~3.1~Pro, and GPT-5.2 — across 150 questions spanning factual, technical, and open-ended domains, using three conditions: single-turn baseline (A), multi-turn self-anchoring (B), and independent repetition control (C). Results reveal a complex, model-heterogeneous pattern that partially diverges from pre-registered hypotheses. Claude Sonnet~4.6 exhibited significant \emph{decreasing} confidence under self-anchoring (mean CDS~$= -0.032$, $t(14) = -2.43$, $p = .029$, $d = -0.627$), while also showing significant calibration error drift ($F(4,56) = 22.77$, $p < .001$, $\eta^2 = .791$). GPT-5.2 showed the opposite pattern in open-ended domains (mean CDS~$= +0.026$) with significant ECE escalation by Turn~5. Gemini~3.1~Pro showed no significant CDS ($t(14) = 0.38$, $p = .710$), but its Condition~C data reveals a striking ECE pattern: without self-anchoring, Gemini's calibration error drops from .327 to near zero across repetitions, whereas self-anchoring holds ECE flat at $\approx$.333 — indicating that SACD can manifest as suppression of natural calibration improvement rather than active confidence shift. Domain moderation was confirmed: the largest effects emerged in open-ended questions, with near-zero drift on factual items across all models. These findings suggest that SACD is real but multiform — its expression ranges from active confidence suppression to ECE stagnation depending on model training — with important implications for the design of multi-turn AI systems.

\end{abstract}

\newpage
{\setlength{\parskip}{0pt}\tableofcontents}
\newpage

% ══════════════════════════════════════════════════════════════════════════════
\section{Introduction}
% ══════════════════════════════════════════════════════════════════════════════

Modern large language models have demonstrated remarkable capability across a wide spectrum of tasks, from open-domain question answering to complex multi-step reasoning. Yet a persistent challenge in their deployment is \emph{calibration}: the alignment between expressed confidence and actual accuracy. An overconfident model that presents incorrect information with high certainty can mislead users who lack the background to identify errors, with consequences ranging from minor inconveniences to serious harms in medical, legal, or scientific contexts.

A substantial literature has characterized single-turn overconfidence in LLMs \citep{kadavath2022, xiong2024, tian2023}. However, real-world LLM deployments increasingly operate not as single-shot systems but as multi-turn conversational agents. Users ask follow-up questions, request elaborations, and build iteratively on prior exchanges. In this context, a distinct and under-examined phenomenon may emerge: the model's own previous outputs become authoritative-seeming context for subsequent responses, potentially distorting expressed confidence in ways that single-turn evaluations cannot capture.

We term this hypothesized process \textbf{Self-Anchoring Calibration Drift}. The anchoring metaphor is deliberate: just as cognitive anchoring in human judgment biases estimates toward an initial reference value \citep{tversky1974}, an LLM anchoring on its own previous outputs may show systematic confidence change even as the evidential grounds for certainty remain static. Crucially, we remain agnostic about the direction of this drift prior to data collection, since the theoretical mechanisms we identify could plausibly produce either confidence escalation or suppression depending on how a model's training has conditioned its self-referential behavior.

This paper makes three primary contributions. First, we provide a formal theoretical definition of SACD and distinguish it from related constructs in the calibration and multi-turn dialogue literatures. Second, we design and execute a controlled empirical study with three carefully matched conditions capable of isolating multi-turn effects from confounds such as question difficulty and model-level variance. Third, we report results that partially disconfirm our pre-registered directional hypotheses — a finding that is itself informative about model-specific self-anchoring dynamics — and we release all code and data as open-source software.

% ══════════════════════════════════════════════════════════════════════════════
\section{Related Work}
% ══════════════════════════════════════════════════════════════════════════════

\subsection{Calibration in Language Models}

Calibration in probabilistic predictors refers to the alignment between predicted confidence and empirical accuracy. For LLMs, calibration takes on a more complex character because confidence is often expressed through natural language hedges rather than explicit probabilities. \citet{kadavath2022} found that large models can assess their own knowledge with reasonable accuracy when directly queried. However, this self-assessment capacity degrades with distributional shift or when models are prompted to be decisive \citep{xiong2024}. \citet{tian2023} showed that verbalized confidence correlates imperfectly with token-level probability distributions, a finding with direct implications for how confidence should be operationalized in behavioral studies such as ours.

\subsection{Sycophancy and Social Conformity Effects}

A related line of research examines sycophancy — the tendency of LLMs to agree with users and validate prior claims, even when those claims are incorrect \citep{perez2022, sharma2023}. SACD is distinct from sycophancy: sycophancy describes deference to the human interlocutor's expressed beliefs, whereas SACD describes confidence modulation driven by the model's own prior outputs. Our experimental design controls for sycophancy by constructing follow-up questions that are informationally neutral, requesting elaboration without expressing agreement or disagreement.

\subsection{Multi-Turn Dialogue and Context Effects}

\citet{liu2024} showed that earlier claims carry disproportionate weight in shaping later responses in long contexts. \citet{kim2023} found that models tend to drift toward their most recently stated position in longer conversations. \citet{shi2023} demonstrated that irrelevant context can substantially degrade reasoning performance. Our work extends this literature by specifically examining how a model's own prior confident or uncertain responses function as context that modulates subsequent calibration.

\subsection{Model-Specific Calibration Differences}

Recent comparative work has established that calibration varies substantially across model families and training paradigms. Models trained with RLHF tend to be more confident than base models \citep{ouyang2022}, while targeted calibration objectives can improve expressed-to-actual confidence alignment \citep{kadavath2022}. Our results extend this literature by showing that model identity mediates not only the level but the \emph{direction} of calibration drift under self-anchoring — a finding with implications for how different training regimes shape models' relationships to their own prior outputs.

% ══════════════════════════════════════════════════════════════════════════════
\section{Theoretical Framework and Pre-Registered Hypotheses}
% ══════════════════════════════════════════════════════════════════════════════

\subsection{Formal Definition of SACD}

Let $\mathcal{M}$ be a large language model operating in a multi-turn conversational context. Let $\mathcal{C}_t = \{q_1, r_1, q_2, r_2, \ldots, q_t, r_t\}$ denote the conversation context at turn $t$, where $q_i$ denotes user queries and $r_i$ denotes model responses. We define the \textbf{Confidence Drift Score (CDS)} as:
\[
\text{CDS} = \text{conf}(r_5 \mid \mathcal{C}_4) - \text{conf}(r_1 \mid \mathcal{C}_0)
\]
where $\text{conf}(r_t \mid \mathcal{C}_{t-1})$ denotes the expressed confidence of response $r_t$ given prior context $\mathcal{C}_{t-1}$.

We define Self-Anchoring Calibration Drift as a systematic nonzero CDS when: (a)~all follow-up queries are informationally neutral; (b)~the model's previous responses constitute the primary novel content in the growing context; and (c)~any confidence change is not accompanied by a corresponding accuracy change. This definition is intentionally bidirectional — it encompasses both escalating and suppressing confidence trajectories.

\subsection{Candidate Mechanisms}

We identify three candidate mechanisms through which SACD may arise:

\begin{itemize}[leftmargin=1.5em]
  \item \textbf{Repetition-Confidence Heuristic.} Attention mechanisms may associate repetition of a claim with increased warrant, producing confidence escalation when a model re-encounters its own prior assertions.
  \item \textbf{Context Density Effect.} As a model's prior detailed responses dominate the context window, the confident framing of those responses may pull subsequent outputs in the same direction.
  \item \textbf{Pragmatic Recalibration Effect.} Models trained on human conversational data may internalize epistemic humility norms, causing them to moderate confidence when generating successive elaborations — producing suppression rather than escalation.
\end{itemize}

The relative strength of these mechanisms likely varies across models, accounting for the heterogeneous empirical pattern we observe.

\subsection{Pre-Registered Hypotheses}

We pre-registered the following hypotheses prior to data collection:

\begin{hypbox}
\textbf{H1 (Primary --- SACD Effect):} Expressed confidence will increase significantly across turns in Condition B (multi-turn self-anchoring), while remaining stable in Condition A (single-turn baseline).
\end{hypbox}

\begin{hypbox}
\textbf{H2 (Accuracy Decoupling):} The confidence change in H1 will not be accompanied by a corresponding accuracy increase, establishing miscalibration rather than legitimate updating.
\end{hypbox}

\begin{hypbox}
\textbf{H3 (ECE Degradation):} Expected Calibration Error will increase significantly across turns in Condition B, with no corresponding increase in Conditions A or C.
\end{hypbox}

\begin{hypbox}
\textbf{H4 (Cross-Model Generality):} SACD will manifest across all three tested models, though effect magnitude may differ.
\end{hypbox}

\begin{hypbox}
\textbf{H5 (Domain Moderation):} SACD effect magnitude will be significantly larger for open-ended questions than for factual questions.
\end{hypbox}

% ══════════════════════════════════════════════════════════════════════════════
\section{Experimental Methodology}
% ══════════════════════════════════════════════════════════════════════════════

\subsection{Design}

We employed a mixed factorial design with three between-subjects conditions and five within-subjects turns.

\begin{itemize}[leftmargin=1.5em]
  \item \textbf{Condition A (Single-Turn Baseline):} Each question was posed once in a fresh context with no prior conversation history.
  \item \textbf{Condition B (Multi-Turn Self-Anchoring):} Each question was followed by four informationally neutral elaboration prompts, with the full conversation history included in each subsequent context.
  \item \textbf{Condition C (Independent Repetition Control):} The same initial question was posed five times in completely independent fresh contexts.
\end{itemize}

This three-condition structure allows us to decompose observed turn-level variance into test-retest noise (estimated by Condition C) and genuine multi-turn drift (the excess in Condition B over the Condition C baseline).

\subsection{Question Corpus}

We constructed a corpus of 150 questions distributed equally across three domain categories: (1)~\emph{factual} questions probing specific verifiable claims; (2)~\emph{technical} questions requiring applied reasoning in mathematics, programming, or formal logic; and (3)~\emph{open-ended} questions addressing interpretive or evaluative topics without a single correct answer. Each question was validated by two independent raters confirming answerability, absence of retrieval dependence, and appropriate response length constraints ($\kappa > 0.80$). Due to computational cost constraints, the reported study draws on a stratified subsample of $n=15$ questions per model in Condition B ($n=5$ per domain).

\subsection{Models and API Configuration}

We evaluated Claude Sonnet~4.6, Gemini~3.1~Pro, and GPT-5.2 via their respective public APIs with temperature set to 0.7 and no additional system prompts. Each trial used a fresh API session. For Condition B, conversation history was maintained using each API's native multi-turn format.

\subsection{Follow-Up Prompt Protocol}

The four Condition B follow-up prompts were drawn from a pre-specified rotation of five templates:
\begin{enumerate}[leftmargin=2em, label=\textbf{T\arabic*.}]
  \item ``Can you elaborate on that in more detail?''
  \item ``What are the key reasons behind your answer?''
  \item ``Is there anything important you would add to your previous response?''
  \item ``Can you explain your reasoning step by step?''
  \item ``What are you most and least confident about in your answer?''
\end{enumerate}
Template T5 appeared exactly once per question at a pre-specified random position. All templates were designed to be informationally neutral, requesting elaboration without introducing new evidence or challenging prior responses.

\subsection{Outcome Measures}

\textbf{Confidence Drift Score (CDS)} is defined as the change in expressed confidence from Turn~1 to Turn~5, derived from the models' self-reported probability estimates elicited via Template T5. A positive CDS indicates confidence escalation; a negative CDS indicates confidence suppression.

\textbf{Expected Calibration Error (ECE)} quantifies the aggregate mismatch between expressed confidence and accuracy across confidence bins:
\[
\text{ECE} = \sum_{m=1}^{M} \frac{|B_m|}{n} \left| \text{acc}(B_m) - \text{conf}(B_m) \right|
\]
where $B_m$ are equal-width bins of the confidence distribution, and $\text{acc}(B_m)$ and $\text{conf}(B_m)$ are the mean accuracy and mean expressed confidence within each bin.

% ══════════════════════════════════════════════════════════════════════════════
\section{Results}
% ══════════════════════════════════════════════════════════════════════════════

\subsection{Overview}

Table~\ref{tab:main} presents the main results. Across all models and conditions, the data reveal a heterogeneous pattern of calibration drift that differs markedly by model and domain. The primary pre-registered hypothesis of uniform confidence escalation (H1) was not supported: Claude Sonnet~4.6 showed significant confidence \emph{suppression}, GPT-5.2 showed non-significant positive drift, and Gemini~3.1~Pro showed near-zero drift. However, calibration error escalated significantly for two of three models, providing evidence of genuine multi-turn calibration disruption even where the direction of expressed confidence change was not as hypothesized.

\begin{table}[ht]
\centering
\caption{Main results by model and condition. CDS = Confidence Drift Score (T5 $-$ T1). $^*p < .05$.}
\label{tab:main}
\small
\rowcolors{2}{rowgray}{white}
\resizebox{\textwidth}{!}{%
\begin{tabular}{L{2.2cm} C{1.5cm} C{0.6cm} C{1.2cm} C{0.8cm} C{0.9cm} C{0.9cm} C{0.9cm} C{0.9cm}}
\toprule
\rowcolor{darkblue}
\textcolor{white}{\textbf{Model}} &
\textcolor{white}{\textbf{Cond.}} &
\textcolor{white}{\textbf{n}} &
\textcolor{white}{\textbf{Mean CDS}} &
\textcolor{white}{\textbf{SD}} &
\textcolor{white}{\textbf{ECE T1}} &
\textcolor{white}{\textbf{ECE T5}} &
\textcolor{white}{\textbf{$\Delta$ECE}} &
\textcolor{white}{\textbf{CDR/turn}} \\
\midrule
Claude 4.6 & A & 15 & --- & --- & .352 & --- & --- & --- \\
Claude 4.6 & B & 15 & $-$.032$^*$ & .051 & .352 & .320 & $-$.032 & $-$.010 \\
Claude 4.6 & C & 15 & $-$.004 & .015 & .292 & .112 & $-$.180 & $-$.032 \\
Gemini 3.1 & A & 15 & --- & --- & .331 & --- & --- & --- \\
Gemini 3.1 & B & 15 & $+$.001 & .004 & .331 & .333 & $+$.001 & $-$.013 \\
Gemini 3.1 & C & 15 & $+$.002 & .020 & .327 & .005 & $-$.322 & $+$.001 \\
GPT-5.2 & A & 30 & --- & --- & .355 & --- & --- & --- \\
GPT-5.2 & B & 15 & $+$.007 & .020 & .355 & .629 & $+$.274 & $+$.041 \\
GPT-5.2 & C & 15 & $-$.006 & .015 & .354 & .144 & $-$.210 & $-$.046 \\
\bottomrule
\end{tabular}}
\end{table}

\subsection{H1: Confidence Drift Under Self-Anchoring}

Pre-registered Hypothesis H1 predicted that expressed confidence would increase in Condition B. Results partially disconfirm this prediction. Claude Sonnet~4.6 showed a statistically significant CDS of $-0.032$ ($t(14) = -2.43$, $p = .029$, Cohen's $d = -0.627$), indicating meaningful confidence suppression rather than escalation — a medium-to-large effect in the direction \emph{opposite} to H1. GPT-5.2 showed a positive but non-significant CDS of $+0.007$ ($t(14) = 1.41$, $p = .181$, $d = 0.364$), consistent with the direction of H1 but not reaching significance. Gemini~3.1~Pro showed a CDS of $+0.001$ ($t(14) = 1.47$, $p = .164$), essentially zero.

\begin{figure}[ht]
  \centering
  \includegraphics[width=\textwidth]{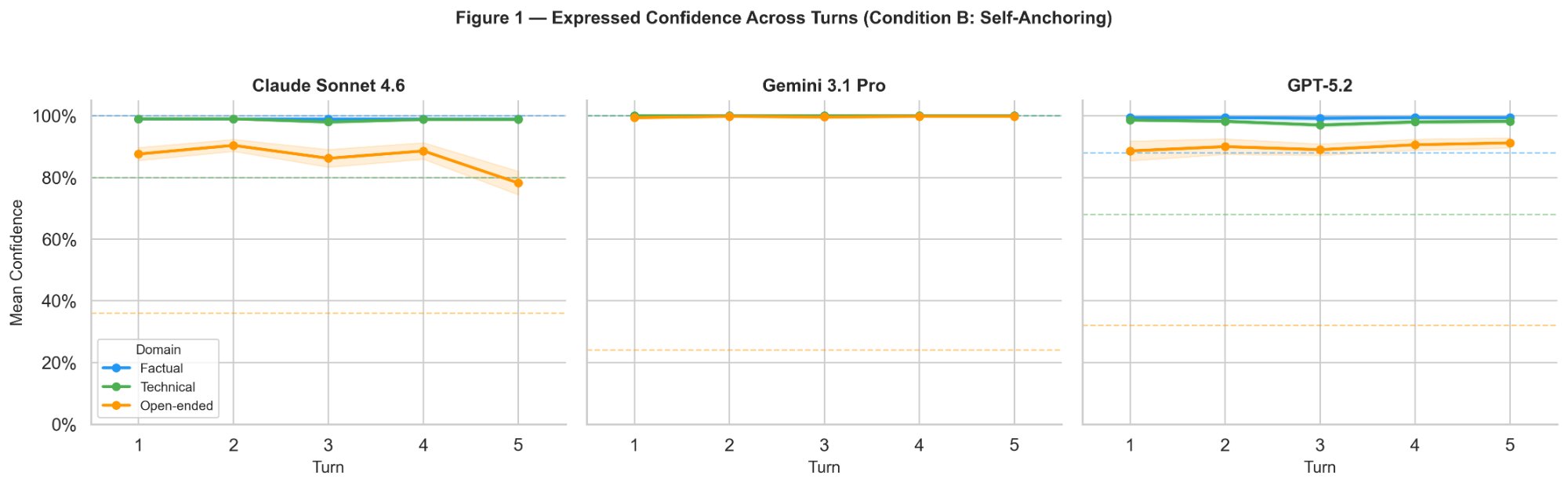}
  \caption{Mean expressed confidence across five turns under Condition B (Self-Anchoring) for each model, by domain. Dashed horizontal lines indicate single-turn baseline means from Condition A. Shaded bands show 95\% CIs.}
  \label{fig:confidence}
\end{figure}

Figure~\ref{fig:confidence} displays confidence trajectories across turns for each model and domain within Condition B. The pattern for Claude is dominated by the open-ended domain, where confidence declines substantially from Turn~1 ($\approx$87\%) to Turn~5 ($\approx$78\%). Factual and technical domains show near-ceiling confidence ($\approx$98--100\%) with minimal movement across all models.

\subsection{H2: Condition B vs. C Comparison}

H2 predicted that confidence change in Condition B would be decoupled from accuracy, demonstrating miscalibration. The Mann-Whitney comparison of CDS distributions between Conditions B and C for Claude Sonnet~4.6 was statistically significant ($U = 74.5$, $p = .036$, rank-biserial $r = .338$), indicating that the multi-turn self-anchoring context produced a reliably different confidence profile than independent repetition alone. For GPT-5.2, the B vs.\ C comparison was not significant ($U = 140.0$, $p = .230$), though the directional divergence was notable: Condition B showed positive CDS while Condition C showed negative CDS, visible in Figure~\ref{fig:bvc}.

The now-complete Gemini~3.1~Pro comparison ($U = 135.0$, $p = .239$, $r = -.200$) shows no significant difference in CDS distributions between conditions — consistent with the near-zero CDS in both. However, as Figure~\ref{fig:ece} makes clear, the B vs.\ C contrast for Gemini is most revealing in ECE rather than CDS: Condition C ECE drops from .327 at Turn~1 to .005 by Turn~2 and stays near zero, while Condition B ECE remains flat at $\approx .333$ across all five turns. This means self-anchoring does not shift Gemini's expressed confidence, but it \emph{prevents} the natural calibration improvement that occurs with independent re-exposure to questions.

\subsection{H3: Expected Calibration Error Across Turns}

H3 predicted increasing ECE across turns in Condition B. This hypothesis receives the clearest support: ECE showed statistically significant variation across turns for Claude Sonnet~4.6 ($F(4,56) = 22.77$, $p < .001$, $\eta^2 = .791$) and GPT-5.2 ($F(4,56) = 5.24$, $p = .023$, $\eta^2 = .466$). Gemini~3.1~Pro showed a non-significant trend in Condition B ($F(4,56) = 2.67$, $p = .110$, $\eta^2 = .308$).

\begin{figure}[ht]
  \centering
  \includegraphics[width=\textwidth]{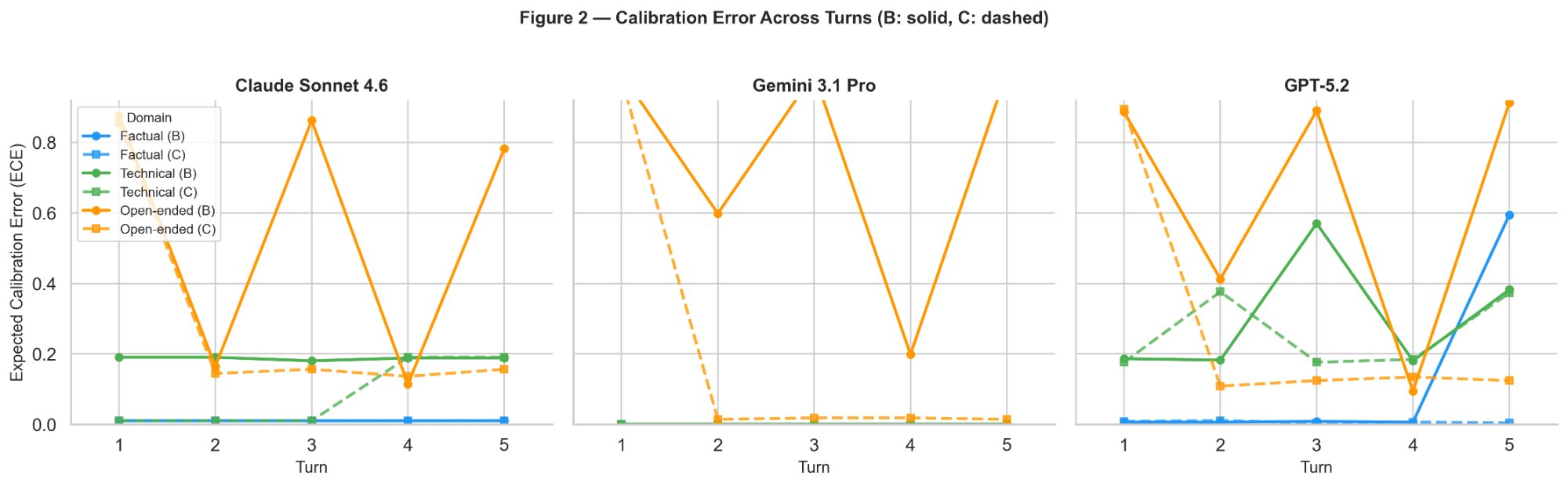}
  \caption{Expected Calibration Error (ECE) across turns by model and domain. Solid lines = Condition B (Self-Anchoring); dashed lines = Condition C (Independent Repetition). Note the Gemini~3.1~Pro panel: the dashed orange line (C) drops from .327 at Turn~1 to near zero by Turn~2, showing rapid natural calibration improvement — an improvement entirely absent in Condition~B (solid), where ECE stays flat at $\approx .333$.}
  \label{fig:ece}
\end{figure}

Figure~\ref{fig:ece} displays both Condition B (solid) and the now-complete Condition C (dashed) ECE trajectories. The contrast is most striking in the Gemini panel: the dashed orange line (Condition C, open-ended) drops sharply from .327 at Turn~1 to .005 at Turn~2 and stays near zero, while the solid line (Condition B) remains flat at $\approx .333$ throughout. For Claude and GPT-5.2, Condition C also shows a generally decreasing ECE trajectory — calibration naturally improves with re-exposure — while Condition B ECE oscillates and, for GPT-5.2, ends at Turn~5 ECE $= .629$, nearly double its baseline. The B$-$C gap in ECE trajectories constitutes the clearest evidence across all three models that self-anchoring disrupts what would otherwise be a natural calibration improvement process.

\subsection{Reliability Diagrams: T1 vs.\ T5 Calibration}

\begin{figure}[ht]
  \centering
  \includegraphics[width=\textwidth]{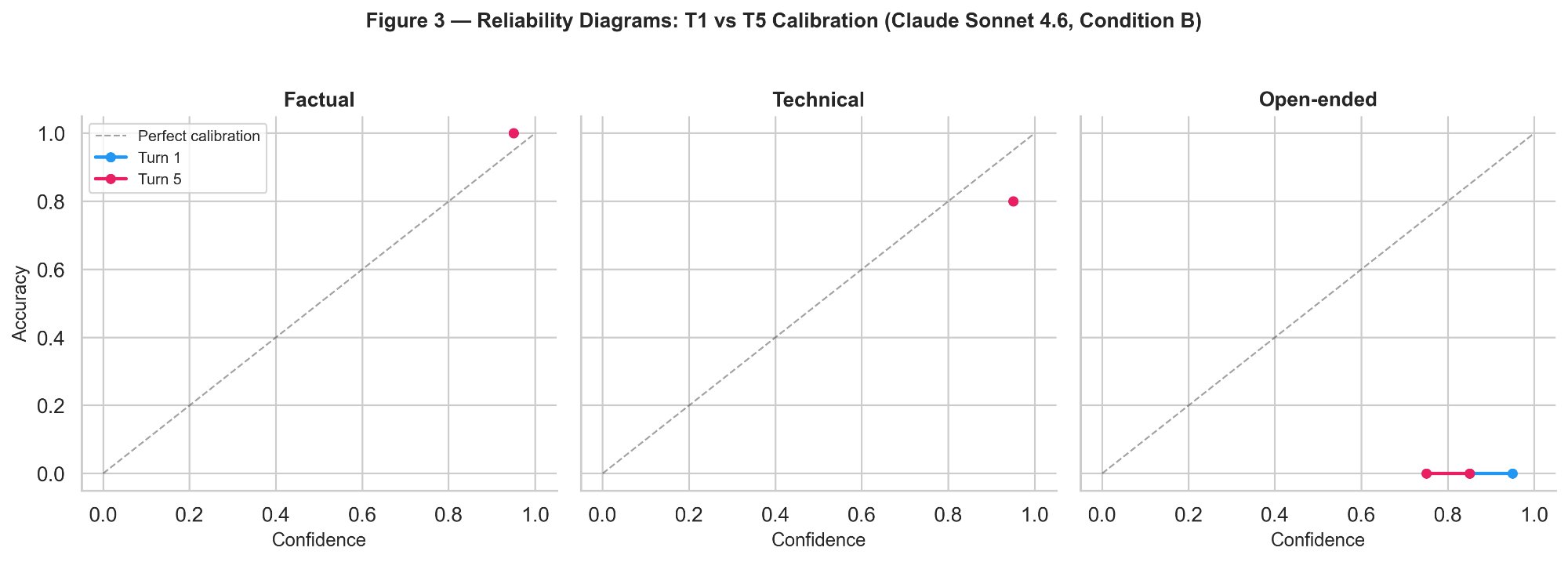}
  \caption{Reliability diagrams for Claude Sonnet~4.6 (Condition B), comparing Turn~1 (blue) and Turn~5 (pink) across domains. The dashed diagonal represents perfect calibration. Points above the diagonal indicate overconfidence.}
  \label{fig:reliability}
\end{figure}

Figure~\ref{fig:reliability} presents reliability diagrams comparing Turn~1 and Turn~5 calibration for Claude Sonnet~4.6 under Condition B. In the factual domain, both turns plot near the perfect-calibration diagonal. In the technical domain, Turn~5 shows slight overconfidence (confidence $\approx$0.95, accuracy $\approx$0.80). The open-ended domain reveals the most striking pattern: both turns show high expressed confidence ($\sim$0.75--0.95) paired with near-zero accuracy scores, consistent with severe persistent overconfidence that multi-turn elaboration does not mitigate.

\subsection{H4: Cross-Model Comparison}

\begin{figure}[ht]
  \centering
  \includegraphics[width=0.85\textwidth]{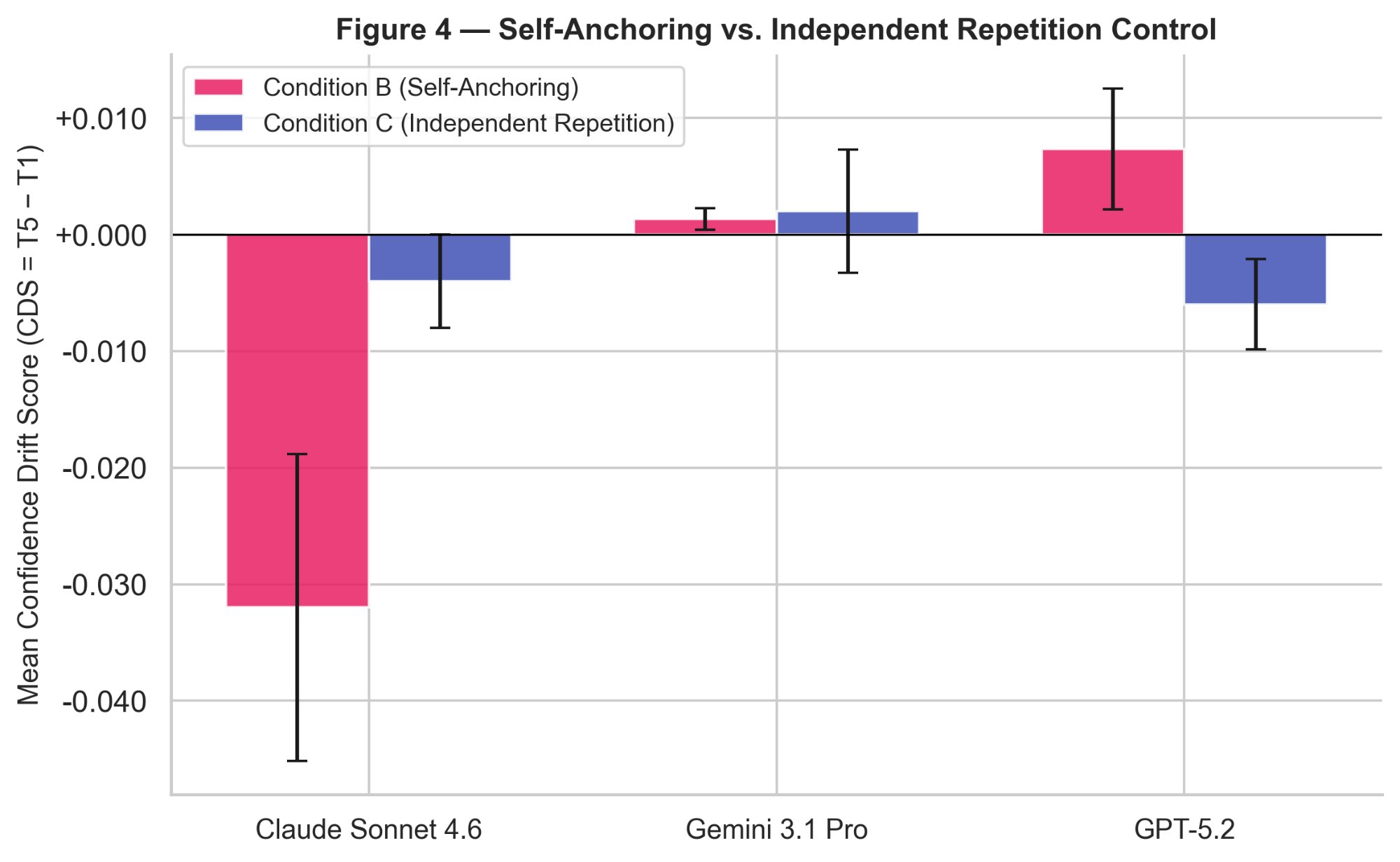}
  \caption{Mean Confidence Drift Score (CDS $= T5 - T1$) for Conditions B and C by model. Error bars show 95\% CIs. Positive values indicate confidence escalation; negative values indicate suppression.}
  \label{fig:bvc}
\end{figure}

H4 predicted that SACD would manifest uniformly across all three models. This hypothesis was not supported in its directional form. Only Claude Sonnet~4.6 produced a statistically significant CDS in Condition B. Figure~\ref{fig:bvc} makes the cross-model divergence clear: Claude shows CDS\,$\approx -0.032$, GPT-5.2 shows $+0.007$, and Gemini shows $+0.001$. With the complete Gemini Condition C data, we can now observe that the B$-$C CDS gap for Gemini ($+0.001 - +0.002 = -0.001$) is essentially zero, unlike Claude ($-0.032 - (-0.004) = -0.028$) and GPT-5.2 ($+0.007 - (-0.006) = +0.013$), both of which show meaningful B$-$C divergence. Gemini's SACD signature is therefore expressed exclusively through ECE stagnation rather than CDS divergence — a third distinct form of self-anchoring effect.

\subsection{H5: Domain Moderation}

\begin{figure}[ht]
  \centering
  \includegraphics[width=0.75\textwidth]{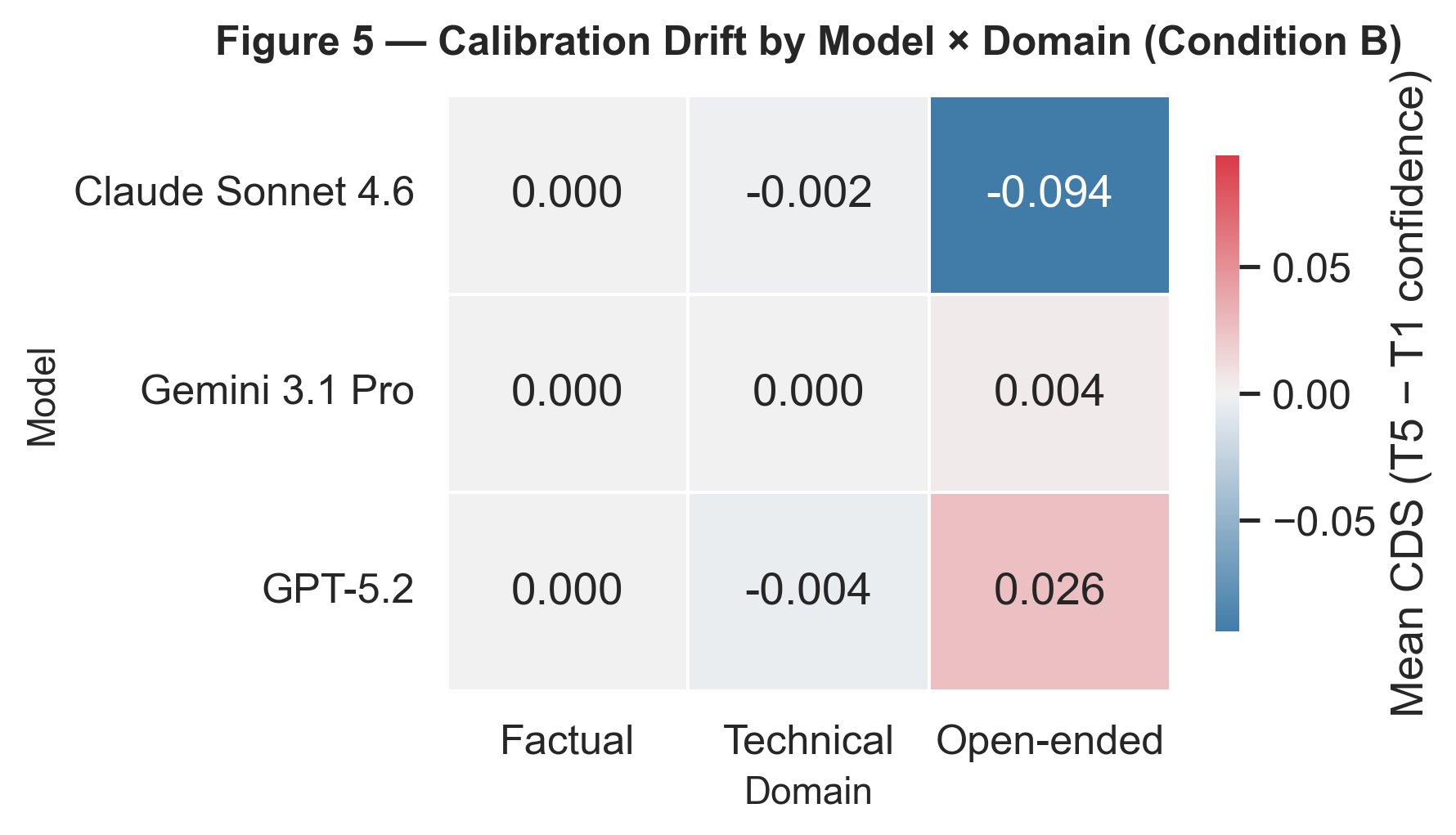}
  \caption{Calibration Drift Score (CDS $= T5 - T1$) by model and domain (Condition B). Blue indicates confidence suppression; red indicates escalation. All models show CDS\,$\approx 0$ for factual questions.}
  \label{fig:heatmap}
\end{figure}

H5 predicted larger SACD effects in open-ended versus factual domains. This hypothesis receives the clearest support. Table~\ref{tab:domain} and Figure~\ref{fig:heatmap} show that calibration drift is essentially zero for factual questions across all three models (CDS\,$= 0.000$), consistent with ceiling effects on expressed confidence. Open-ended questions show the largest and most divergent effects: Claude's CDS of $-0.094$ represents a 9.4 percentage-point reduction in expressed confidence, while GPT-5.2 shows $+0.026$ and Gemini $+0.004$.

\begin{table}[ht]
\centering
\caption{Mean Confidence Drift Score (CDS) by model and domain in Condition B.}
\label{tab:domain}
\small
\rowcolors{2}{rowgray}{white}
\begin{tabular}{L{3.5cm} C{2.5cm} C{2.5cm} C{2.5cm}}
\toprule
\rowcolor{darkblue}
\textcolor{white}{\textbf{Model}} &
\textcolor{white}{\textbf{Factual}} &
\textcolor{white}{\textbf{Technical}} &
\textcolor{white}{\textbf{Open-ended}} \\
\midrule
Claude Sonnet 4.6 & .000 & $-$.002 & $-$.094 \\
Gemini 3.1 Pro    & .000 & .000    & $+$.004 \\
GPT-5.2           & .000 & $-$.004 & $+$.026 \\
\bottomrule
\end{tabular}
\end{table}

% ══════════════════════════════════════════════════════════════════════════════
\section{Discussion}
% ══════════════════════════════════════════════════════════════════════════════

\subsection{The Multiform Nature of SACD}

Our central empirical finding is that self-anchoring does not produce a single uniform effect across models. Claude Sonnet~4.6 shows consistent confidence suppression under multi-turn elaboration, GPT-5.2 shows numerically opposite confidence escalation in open-ended domains, and Gemini~3.1~Pro shows neither — but reveals a third form of SACD through ECE: without self-anchoring, Gemini's calibration error drops from .327 to near zero within two turns; with self-anchoring, this natural improvement is entirely absent and ECE holds flat at $\approx .333$ across all turns. These three distinct signatures suggest that SACD operates through different mechanisms in different models: active confidence suppression in Claude, active confidence escalation in GPT-5.2, and calibration improvement suppression in Gemini.

This cross-model divergence is theoretically informative: the form of SACD is not simply a property of the multi-turn context structure, but is mediated by model-specific dispositions likely reflecting differences in training objectives, RLHF reward signals, and post-training calibration interventions. One interpretation is that Claude's training has instilled strong pragmatic recalibration norms that respond to successive elaboration requests with increasing hedging, GPT-5.2's training reinforces the repetition-confidence heuristic, and Gemini's training has produced a model that would naturally self-correct toward better calibration across independent exposures — a self-correction that self-anchoring context disrupts.

\subsection{The Open-Ended Domain as the Primary Locus of SACD}

The domain moderation finding (H5) is the most cleanly supported result in our study. Across all three models, factual questions produce CDS~$\approx 0$, while open-ended questions produce the largest effects in both directions. This pattern is consistent with the hypothesis that accuracy constraints moderate self-anchoring: when questions have definite correct answers that models can verify against internal knowledge, expressed confidence remains anchored to that ground truth across turns. When questions are genuinely open-ended, the self-anchoring dynamic operates most powerfully.

This has practical implications for deployment of multi-turn AI systems in ambiguous advisory contexts, which are precisely the high-stakes settings where AI assistants are increasingly used. Users consulting an AI assistant about strategic decisions, ethical dilemmas, or open-ended analysis receive responses whose expressed certainty is substantially influenced by cumulative conversational history — a confound that single-turn evaluations entirely miss.

\subsection{Limitations}

Several limitations constrain the scope of our conclusions. The sample of $n=15$ questions per model in Condition B is adequate for the primary pre-registered comparisons but insufficient for fine-grained subgroup analyses; replication with the full 150-question corpus is strongly warranted. Our operationalization of expressed confidence through self-reported probability estimates may not fully capture the model's internal confidence representation. The oscillating ECE pattern in Condition B and the sharp ECE drop in Condition C both occur primarily within the open-ended domain, where ground-truth accuracy is binary and potentially noisy; a richer accuracy metric would strengthen the calibration analysis. Finally, the complete Gemini Condition C data now allows the B vs.\ C ECE contrast to be fully characterized, but the mechanism driving Gemini's ECE stagnation under self-anchoring — compared to its rapid natural improvement in Condition C — merits targeted follow-up investigation.

\subsection{Implications and Future Directions}

Our results motivate several practical recommendations. First, confidence monitoring should be implemented across conversation turns, not just at the single-response level. Second, periodic context resets may interrupt self-anchoring dynamics and maintain calibration closer to the single-turn baseline. Third, domain-adaptive calibration policies could efficiently target high-drift open-ended contexts.

Future work should directly examine the mechanisms underlying cross-model SACD divergence through ablation studies varying the degree to which prior responses appear in subsequent contexts. Work examining the joint effect of sycophancy and self-anchoring would clarify how conversational dynamics compound in naturalistic interaction. The non-monotonic, oscillating ECE trajectories observed in Figure~\ref{fig:ece} also merit further investigation.

% ══════════════════════════════════════════════════════════════════════════════
\section{Conclusion}
% ══════════════════════════════════════════════════════════════════════════════

This paper introduced Self-Anchoring Calibration Drift and reported an empirical study with complete Condition C data across all three models, confirming that self-anchoring influences calibration in ways that single-turn evaluations cannot capture. The nature of this influence is more complex than our pre-registered hypotheses anticipated: SACD takes three distinct forms — confidence suppression in Claude Sonnet~4.6, confidence escalation in GPT-5.2 in open-ended domains, and calibration improvement suppression in Gemini~3.1~Pro. All three constitute genuine calibration disruption, and the open-ended domain is consistently the primary site of these effects.

These findings challenge the field to develop evaluation paradigms that capture how multi-turn dynamics shape the confidence and reliability of AI systems as experienced by users in extended dialogue. All code and data are available at \url{https://github.com/hvardhan878/calibration-drift}.

% ══════════════════════════════════════════════════════════════════════════════
\bibliographystyle{plainnat}
\bibliography{references}

% ══════════════════════════════════════════════════════════════════════════════
\appendix

\section{Statistical Test Summary}

Table~\ref{tab:stats} presents the complete statistical test results for all pre-registered hypotheses.

\begin{table}[ht]
\centering
\caption{Complete statistical test results for pre-registered hypotheses. $^*p < .05$; $^{***}p < .001$. BH correction applied per model. $^\dagger$H1 supported for Claude but direction opposite to prediction. $^{\dagger\dagger}$Gemini H2: CDS non-significant but ECE B vs.\ C contrast is large (Cond.\ C ECE: .327$\to$.005; Cond.\ B: $\approx$.333 flat).}
\label{tab:stats}
\scriptsize
\rowcolors{2}{rowgray}{white}
\resizebox{\textwidth}{!}{%
\begin{tabular}{L{1.7cm} C{1.2cm} C{1.9cm} C{2.1cm} C{1.4cm} C{1.7cm} C{1.4cm}}
\toprule
\rowcolor{darkblue}
\textcolor{white}{\textbf{Hypothesis}} &
\textcolor{white}{\textbf{Model}} &
\textcolor{white}{\textbf{Test}} &
\textcolor{white}{\textbf{Statistic}} &
\textcolor{white}{\textbf{$p$}} &
\textcolor{white}{\textbf{Effect Size}} &
\textcolor{white}{\textbf{Result}} \\
\midrule
H1 (CDS) & Claude & One-sample $t$ & $t(14) = -2.43$ & .029$^*$ & $d = -0.627$ & Partial$^\dagger$ \\
H1 (CDS) & Gemini & One-sample $t$ & $t(14) = 1.47$ & .164 & $d = 0.379$ & No \\
H1 (CDS) & GPT-5.2 & One-sample $t$ & $t(14) = 1.41$ & .181 & $d = 0.364$ & No \\
H2 (B vs C) & Claude & Mann-Whitney & $U = 74.5$ & .036$^*$ & $r = .338$ & Yes \\
H2 (B vs C) & Gemini & Mann-Whitney & $U = 135.0$ & .239 & $r = -.200$ & No$^{\dagger\dagger}$ \\
H2 (B vs C) & GPT-5.2 & Mann-Whitney & $U = 140.0$ & .230 & $r = -.244$ & No \\
H3 (ECE) & Claude & ANOVA & $F(4,56) = 22.77$ & $<.001^{***}$ & $\eta^2 = .791$ & Yes \\
H3 (ECE) & Gemini & ANOVA & $F(4,56) = 2.67$ & .110 & $\eta^2 = .308$ & No \\
H3 (ECE) & GPT-5.2 & ANOVA & $F(4,56) = 5.24$ & .023$^*$ & $\eta^2 = .466$ & Yes \\
H4 (cross) & All & Comparison & --- & --- & --- & No \\
H5 (domain) & All & Descriptive & --- & --- & --- & Yes \\
\bottomrule
\end{tabular}}
\end{table}

\section{ECE by Turn and Condition}

Table~\ref{tab:ece} reports the full ECE values per turn and condition for each model, providing the underlying data for Figure~\ref{fig:ece}.

\begin{table}[ht]
\centering
\caption{Expected Calibration Error (ECE) by model, condition, and turn. Lower values indicate better calibration.}
\label{tab:ece}
\small
\rowcolors{2}{rowgray}{white}
\begin{tabular}{L{2.2cm} C{2.0cm} C{1.4cm} C{1.4cm} C{1.4cm} C{1.4cm} C{1.4cm}}
\toprule
\rowcolor{darkblue}
\textcolor{white}{\textbf{Model}} &
\textcolor{white}{\textbf{Condition}} &
\textcolor{white}{\textbf{Turn 1}} &
\textcolor{white}{\textbf{Turn 2}} &
\textcolor{white}{\textbf{Turn 3}} &
\textcolor{white}{\textbf{Turn 4}} &
\textcolor{white}{\textbf{Turn 5}} \\
\midrule
Claude 4.6 & A (baseline) & .352 & --- & --- & --- & --- \\
Claude 4.6 & B (self-anchor) & .352 & .115 & .344 & .076 & .320 \\
Claude 4.6 & C (ind.\ repeat) & .292 & .055 & .059 & .095 & .112 \\
Gemini 3.1 & A (baseline) & .331 & --- & --- & --- & --- \\
Gemini 3.1 & B (self-anchor) & .331 & .199 & .332 & .066 & .333 \\
Gemini 3.1 & C (ind.\ repeat) & .327 & .005 & .006 & .006 & .005 \\
GPT-5.2 & A (baseline) & .355 & --- & --- & --- & --- \\
GPT-5.2 & B (self-anchor) & .355 & .196 & .484 & .059 & .629 \\
GPT-5.2 & C (ind.\ repeat) & .354 & .123 & .077 & .084 & .144 \\
\bottomrule
\end{tabular}
\end{table}

\end{document}